\documentclass[10pt,twocolumn,letterpaper]{article}

\usepackage{iccv}
\usepackage{times}
\usepackage{epsfig}
\usepackage{graphicx}
\usepackage{amsmath}
\usepackage{amssymb}
\usepackage{subcaption}

\usepackage{multirow}
\usepackage{appendix}
\usepackage{tabularx}
\usepackage{enumitem}
\usepackage{csquotes}

\usepackage[pagebackref=true,breaklinks=true,letterpaper=true,colorlinks,bookmarks=false]{hyperref}

\iccvfinalcopy % *** Uncomment this line for the final submission

\def\iccvPaperID{9466} % *** Enter the ICCV Paper ID here

\newcommand{\smallsec}[1]{\vspace{0.05in} \noindent {\bf #1.}}

\newcommand{\localqa}{{\bf PointQA-Local} }
\newcommand{\looktwiceqa}{{\bf PointQA-LookTwice} }
\newcommand{\generalqa}{{\bf PointQA-General} }

\newcommand{\localqan}{{\bf PointQA-Local}}
\newcommand{\looktwiceqan}{{\bf PointQA-LookTwice}}

\newcommand{\localmodel}{TBD:PointQA-Local-model }

\newcommand{\segmmodel}{TBD:PointQA-Local-Segm-model }

\newcommand{\localmodeln}{TBD:PointQA-Local-model}

\newcommand{\segmmodeln}{TBD:PointQA-Local-Segm-model}

\makeatletter
\let\@fnsymbol\@arabic
\makeatother

\ificcvfinal\fi

\begin{document}

%%%%%%%%% TITLE
\title{Point and Ask: Incorporating Pointing into Visual Question Answering}

% Exploring and addressing spatial ambiguity in VQA
% Exploring and addressing 
% reference, spatial grounding, pointing, 
% Point and ask: Exploring point-based spatial grounding in Visual Question Answering
% Point and ask: Exploring point input in Visual Question Answering
% Point and ask: Utilizing point input in Visual Question Answering
% Point and ask: Visual Question Answering with a point input
% Point and ask: Visual Question Answering requiring a point input
% Point and ask: Visual Question Answering with pointing
% Point and ask: Pointing-based Visual Question Answering
% Point and ask: Exploring pointing-driven Visual Question Answering
% Point and Ask: Incorporating Pointing into Visual Question Answering
% Exploring pointing in VQA
% 

% Authors: Arjun, Nobline, Will, Olga

\author{Arjun Mani\thanks{AM is currently a PhD student at Columbia University. Work was conducted while he was an undergraduate student at Princeton University. Columbia email address is provided above for contact regarding this work.} , Nobline Yoo, Will Hinthorn\thanks{WH is currently at Robust Intelligence. Work was conducted while he was an undergraduate student at Princeton University.} , Olga Russakovsky \\
Princeton University\\
{\tt\small asm2290@columbia.edu, olgarus@cs.princeton.edu}}
% \and
% Will Hinthorn\\
% % First line of institution2 address\\
% % {\tt\small secondauthor@i2.org}

% \and
% Nobline Yoo\\
% % First line of institution2 address\\
% % {\tt\small secondauthor@i2.org}

% \and
% Olga Russakovsky\\
% {\tt\small \{arjuns, olgarus\}@cs.princeton.edu}

\maketitle
\ificcvfinal\thispagestyle{empty}\fi

%%%%%%%%%%%%%%%%%%%%%%%%%%%%%%%%%%%%%%%%%%%%%%%%%%%%%
%%%%%%%%%%%%%%%%%%%%%%%%%%%%%%%%%%%%%%%%%%%%%%%%%%%%%
%%%%%%%%%%%%%%%%%%%%%%%%%%%%%%%%%%%%%%%%%%%%%%%%%%%%%
%%%%%%%%%%%%%%%%%%%%%%%%%%%%%%%%%%%%%%%%%%%%%%%%%%%%%
%%%%%%%
%%%%%%% ABSTRACT
%%%%%%%
%%%%%%%%%%%%%%%%%%%%%%%%%%%%%%%%%%%%%%%%%%%%%%%%%%%%%
%%%%%%%%%%%%%%%%%%%%%%%%%%%%%%%%%%%%%%%%%%%%%%%%%%%%%
%%%%%%%%%%%%%%%%%%%%%%%%%%%%%%%%%%%%%%%%%%%%%%%%%%%%%
%%%%%%%%%%%%%%%%%%%%%%%%%%%%%%%%%%%%%%%%%%%%%%%%%%%%%

\begin{abstract}
Visual Question Answering (VQA) has become one of the key benchmarks of visual recognition progress. Multiple VQA extensions have been explored to better simulate real-world settings: different question formulations, changing training and test distributions, conversational consistency in dialogues, and explanation-based answering. In this work, we further expand this space by considering visual questions that include a spatial point of reference. Pointing is a nearly universal gesture among humans, and real-world VQA is likely to involve a gesture towards the target region. 

%In this work, we further expand this space by considering the nearly-universal gesture of pointing, and explore visual questions that further contain a spatial point of reference. 

Concretely, we (1) introduce and motivate point-input questions as an extension of VQA, (2) define three novel classes of questions within this space, and (3) for each class, introduce both a benchmark dataset and a series of model designs to handle its unique challenges. There are two key distinctions from prior work. First, we explicitly design the benchmarks to require the point input, i.e., we ensure that the visual question cannot be answered accurately without the spatial reference. Second, we explicitly explore the more realistic point spatial input rather than the standard but unnatural bounding box input. Through our exploration we uncover and address several visual recognition challenges, including the ability to reason both locally and globally about the image, and to effectively combine visual, language and spatial inputs. Code is available at: \href{https://github.com/princetonvisualai/pointingqa}{\emph{github.com/princetonvisualai/pointingqa}}.

\end{abstract}

%%%%%%%%%%%%%%%%%%%%%%%%%%%%%%%%%%%%%%%%%%%%%%%%%%%%%
%%%%%%%%%%%%%%%%%%%%%%%%%%%%%%%%%%%%%%%%%%%%%%%%%%%%%
%%%%%%%%%%%%%%%%%%%%%%%%%%%%%%%%%%%%%%%%%%%%%%%%%%%%%
%%%%%%%%%%%%%%%%%%%%%%%%%%%%%%%%%%%%%%%%%%%%%%%%%%%%%
%%%%%%%
%%%%%%% INTRO
%%%%%%%
%%%%%%%%%%%%%%%%%%%%%%%%%%%%%%%%%%%%%%%%%%%%%%%%%%%%%
%%%%%%%%%%%%%%%%%%%%%%%%%%%%%%%%%%%%%%%%%%%%%%%%%%%%%
%%%%%%%%%%%%%%%%%%%%%%%%%%%%%%%%%%%%%%%%%%%%%%%%%%%%%
%%%%%%%%%%%%%%%%%%%%%%%%%%%%%%%%%%%%%%%%%%%%%%%%%%%%%

\section{Introduction}

Visual Question Answering (VQA) has emerged as a popular and challenging task in computer vision~\cite{antol2015vqa,goyal2019making,zhu2016visual7w,hudson2019gqa,visualdialog,embodiedqa}. When the task was first introduced, the challenge of answering a natural language question about an image was already a significant leap beyond more conventional tasks such as object recognition. Since then, questions in VQA benchmarks have been increasingly growing in complexity: for example, ``Are there any cups to the left of the tray on top of the table?'' is an example question from the recent GQA~\cite{hudson2019gqa}) benchmark. Such questions effectively test the ability of VQA agents to parse very complex sentences, but arguably are becoming less realistic. 
In a real-world setting, it's unlikely that a human would use the phrase ``to the left of the tray on top of the table.'' It's far more likely that they would instead ask ``Are there any cups \emph{over there}'' and point to the left of the tray. In fact, human psychology literature shows that pointing to interesting objects or situations is one of the first ways by which babies communicate intention~\cite{oates2004cognitive,malle2001intentions}. Understanding pointing as part of a visually grounded dialog with humans would naturally be a key ability of real-world AI systems.

%Questions in VQA benchmarks are growing in complexity, in particular through increasingly challenging references: e.g., by asking ``what color is the umbrella \emph{on the right}?'' or . These questions test the \todo{XX}.

%Since then, a number of additional extensions have been developed. For example, Visual Dialog~\cite{XXX} tasks the machine with maintaining a consistent conversation, \todo{XXX tests the ability of the model to shift priors, YYY tests novel concepts..., ZZZ embedded VQA}

%Visual7W~\cite{zhu2016visual7w} includes questions about specific image regions, and GQA~\cite{hudson2019gqa} further requires models to attend to the right image region in addition to providing the right answer. 

%while Visual7W~\cite{zhu2016visual7w} or GQA~\cite{hudson2019gqa} include spatial grounding input to illustrate which information is necessary and relevant to answer a question.

%However, one of the most natural ways for humans to refer to an object is actually by pointing: ``That cat over there'' (point) or ``What is that over there?'' (point). Pointing to interesting objects or situations is one of the first ways by which babies communicate intention. In fact, the development of \emph{shared attention}, where one individual is able to divert another's attention to an object e.g., by pointing, is an important step of cognitive development in children~\cite{oates2004cognitive,malle2001intentions}. 
 
%make new figure (real-world - smth not captured in local/looktwice)
 \begin{figure}[t]
\includegraphics[width=\linewidth]{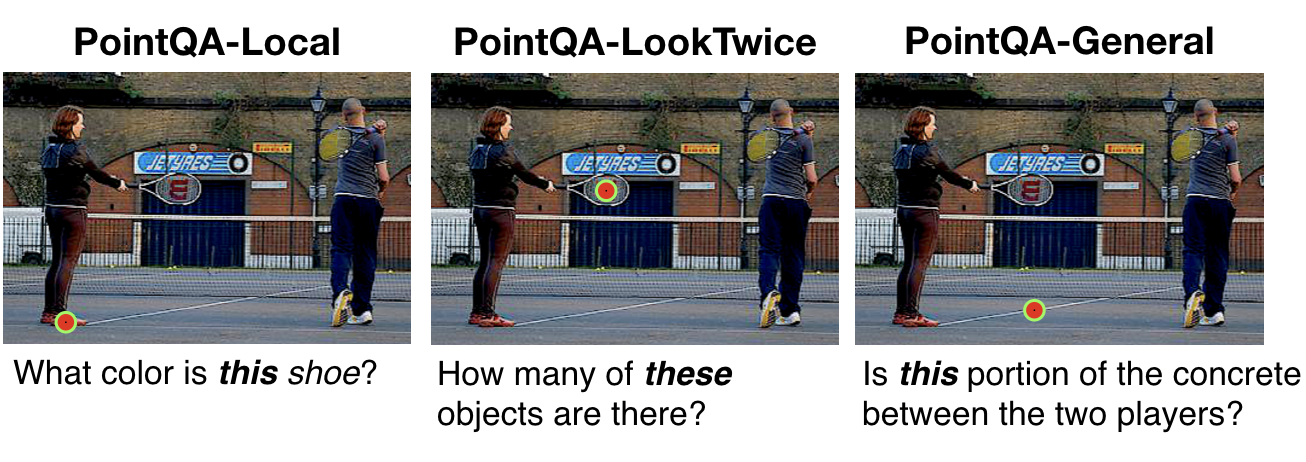}
\caption{Three types of visual questions requiring a point input (red) which we propose and analyze in this work.}
\label{fig:pullfig}
\end{figure}
 
We thus propose to expand the space of VQA by considering visual questions that further include a spatial point of reference for context. Prior works have used visual grounding to expand the question space of VQA: e.g., Visual7W~\cite{zhu2016visual7w} introduced ``which'' question with image regions as candidate answers; Visual Genome~\cite{krishna2017visual} contains questions that are associated with particular regions in the image; GQA~\cite{hudson2019gqa} leverages the grounded scene graph in its question construction process. There are two key distinctions of our proposal from this line of work. First, we explicitly design the benchmarks to \emph{require} the point input, i.e., we ensure that the visual question cannot be answered accurately without the spatial reference. Second, we explore the more realistic \emph{point} spatial input rather than the standard but unnatural bounding box used in~\cite{zhu2016visual7w,krishna2017visual,hudson2019gqa}. %Within computer vision, pointing to objects~\cite{Bearman16,clickcarving,Papadopoulos17} and materials~\cite{bell15minc} has been used as a form of cost-effective supervision but to the best of our knowledge never examined in depth beyond cost-accuracy tradeoffs.

We introduce a set of tasks exploring different aspects and challenges of point-based spatial disambiguation, shown in Fig.~\ref{fig:pullfig}. In all tasks the input is an image and a single pixel in the image corresponding to the spatial grounding. The target output is a multiple choice answer.

We first consider two narrow settings: (1) \localqan, where only the local region around the point is relevant to the question, e.g., ``What color is \emph{this}  shirt?" where \emph{this} is specified by a pixel in the image (Sec.~\ref{sec:localqa}) and (2) \looktwiceqan, which requires a global understanding of the image, e.g., ``is another shirt \emph{this} color?'' (Sec.~\ref{sec:looktwiceqa}). For each we construct a corresponding dataset from Visual Genome annotations~\cite{krishna2017visual}, with 57,628 questions across 18,830 images and 57,405 questions across 34,676 images, respectively. We then modify and benchmark VQA models~\cite{pythia18arxiv,jiang2020defense} to  incorporate the point input.

Finally, we consider the general setting of unconstrained questions which require point disambiguation in \generalqa by adapting human-written questions from the Visual7W~\cite{zhu2016visual7w} dataset, resulting in 319,300 questions over 25,420 images (Sec.~\ref{sec:generalqa}). We modify the state-of-the-art Pythia~\cite{pythia18arxiv}, MCAN~\cite{yu2019mcan} and LXMERT~\cite{tan2019lxmert} models to incorporate the point input, and demonstrate their effectiveness in this new setting.

To summarize, our work advances VQA along a new dimension. Concretely, we (1) introduce and motivate point-input questions as an extension of VQA, (2) design a set of benchmark datasets, and (3) introduce effective model extensions to handle the unique challenges of this space.

%Our aim is to provide a clear roadmap for future innovations in this space. 

%We empirically demonstrate that our point-aware models outperform existing VQA baselines on pointing QA, laying the groundwork for future innovations in this space.

%We explore several methods to incorporate a point and show that our proposed model can successfully answer Local-QA questions; moreover, we benchmark it on Intent-QA and show that it can infer the intended object of reference. For LookTwice-QA, our proposed model can perform both global and local reasoning when incorporating the point.

% Not really satisfied with this paragraph, so hope we can rephrase here.

% \todo{Would be great to summarize a few key takeaways from the results}
% 

%%%%%%%%%%%%%%%%%%%%%%%%%%%%%%%%%%%%%%%%%%%%%%%%%%%%%
%%%%%%%%%%%%%%%%%%%%%%%%%%%%%%%%%%%%%%%%%%%%%%%%%%%%%
%%%%%%%%%%%%%%%%%%%%%%%%%%%%%%%%%%%%%%%%%%%%%%%%%%%%%
%%%%%%%%%%%%%%%%%%%%%%%%%%%%%%%%%%%%%%%%%%%%%%%%%%%%%
%%%%%%%
%%%%%%% RELATED WORK
%%%%%%%
%%%%%%%%%%%%%%%%%%%%%%%%%%%%%%%%%%%%%%%%%%%%%%%%%%%%%
%%%%%%%%%%%%%%%%%%%%%%%%%%%%%%%%%%%%%%%%%%%%%%%%%%%%%
%%%%%%%%%%%%%%%%%%%%%%%%%%%%%%%%%%%%%%%%%%%%%%%%%%%%%
%%%%%%%%%%%%%%%%%%%%%%%%%%%%%%%%%%%%%%%%%%%%%%%%%%%%%

\section{Related Work}

%\subsection{Spatial Grounding in VQA} 

\smallsec{Spatial Grounding in VQA} Visual grounding has become a central idea in the VQA community~\cite{zhu2016visual7w,goyal2019making,agrawal2018dont,hudson2019gqa}. It is increasingly seen as important that VQA models localize the object being asked about to answer a visual question. This idea has influenced the development of several datasets, including Visual7W \cite{zhu2016visual7w} and GQA \cite{hudson2019gqa}. In these datasets, bounding boxes are provided for each object mentioned in the question or answer. To encourage grounding and counteract language priors, \cite{goyal2019making} introduced the VQA 2.0 Dataset in 2017, which consists of complementary images for each question. 
\cite{agrawal2018dont} further introduce the VQA-CP dataset, where priors differ in training and test splits. All these works indicate the importance of visual grounding for the VQA task.

A few works have used visual grounding to expand the question space of VQA. The authors of Visual7W introduce a \textit{pointing QA} task, involving a `which' question and image regions as candidate answers \cite{zhu2016visual7w}. They use their manually collected object-level groundings to construct such questions. A number of works have also explored grounding in the context of embodied question answering, which requires an agent to explore an environment to answer a visual question \cite{embodiedqa, interactiveqa}. Other grounding-based questions include region-based QAs in Visual Genome \cite{krishna2017visual}, where the question is associated with a particular region of the image. However, it is not necessary for answering the question that the region be provided: we sampled 100 questions randomly and only 17\% actually required the region to produce the correct answer. To our knowledge, we are the first to introduce a benchmark where a spatial grounding signal is explicitly required to answer a question.

% the agent is asked a question in a 3D simulated environment. Answering the question requires intelligently exploring the space to locate the region relevant to the question, effectively grounding the question in the specific location in the environment.

The importance of visual grounding has also influenced the development of VQA models. In particular, most state-of-the-art VQA models have an attention mechanism on the image \cite{tan2019lxmert,vilbert,pythia18arxiv,yu2019mcan}. The relative success of these models indicates the importance of successful visual grounding in the VQA task. Several works include pixel-wise prediction as a primary or auxiliary task of the VQA model. \cite{zhang2019interpretable} mine ground-truth attention maps from Visual Genome and include attention prediction explicitly as an auxiliary task of the model. Other works output a visual justification for the answer as a heatmap \cite{park2018multimodal} or a semantic segmentation of the visual entities relevant to the question \cite{gan2017vqs}. However, the challenge of actually accepting a spatial grounding input into VQA models has not been previously explored.

% To our knowledge, no previous work has explicitly predicted over the answer distribution at each spatial location in the image.

%\subsection{Point Input} 

\smallsec{Point input} Despite being ubiquitous for humans, pointing as a way of communicating intention has been underexplored in computer vision. Studies in the robotics~\cite{hild2003object,Sauppe2014robot,raza2013human} or human-computer interaction~\cite{merrill2007augmenting} communities have largely been limited to simple, constrained environments. In vision, pointing has been used as a form of cost-effective supervision~\cite{bearman2016whats,jain2016click,Papadopoulos2014training,bell2015material} but never examined in depth beyond cost-accuracy tradeoffs. The use of pointing as a communicative gesture in humans ~\cite{oates2004cognitive,malle2001intentions}  motivates deeper study in a computer vision context.

%%%%%%%%%%%%%%%%%%%%%%%%%%%%%%%%%%%%%%%%%%%%%%%%%%%%%
%%%%%%%%%%%%%%%%%%%%%%%%%%%%%%%%%%%%%%%%%%%%%%%%%%%%%
%%%%%%%%%%%%%%%%%%%%%%%%%%%%%%%%%%%%%%%%%%%%%%%%%%%%%
%%%%%%%%%%%%%%%%%%%%%%%%%%%%%%%%%%%%%%%%%%%%%%%%%%%%%
%%%%%%%
%%%%%%% LOCAL-QA
%%%%%%%
%%%%%%%%%%%%%%%%%%%%%%%%%%%%%%%%%%%%%%%%%%%%%%%%%%%%%
%%%%%%%%%%%%%%%%%%%%%%%%%%%%%%%%%%%%%%%%%%%%%%%%%%%%%
%%%%%%%%%%%%%%%%%%%%%%%%%%%%%%%%%%%%%%%%%%%%%%%%%%%%%
%%%%%%%%%%%%%%%%%%%%%%%%%%%%%%%%%%%%%%%%%%%%%%%%%%%%%

\section{\localqan: reasoning about a region}
\label{sec:localqa}
% \olga{todo: update}
We now begin our exploration of the space of pointing questions, starting with a simpler \localqa setting: questions involves queries about the attributes of a particular object (e.g. ``What color is \emph{this} car?"). A local region around the point is completely sufficient to answer the question. In Sec.~\ref{sec:localqa-data} we construct a corresponding dataset using existing annotations from Visual Genome~\cite{krishna2017visual}. We design the benchmark such that a point is \emph{required} to answer the question, ensuring that the image as a whole contains multiple possible answers.  

\localqa allows us to investigate the capabilities of the model to accept a point input, without additional complexities introduced in Sections~\ref{sec:looktwiceqa} and \ref{sec:generalqa}. Importantly,  in Sec.~\ref{sec:localqa-results} we demonstrate that the model is able to understand the question and the task enough to vary attention around the point as needed (consider ``What color is this shirt?" versus ``What action is this person doing?").

\subsection{\localqa dataset}
\label{sec:localqa-data}

Our \localqa dataset contains questions that require reasoning over an image region, with the point input disambiguating the correct region. We use three templates: %(1) ``What color is this [object class]?'' (2) ``What shape is this [object class]?'' and (3) ``What is this [object class] doing?''
\setlist{nolistsep}
\begin{enumerate}[noitemsep]
    \item What color is \emph{this} [object class]?
    \item What shape is \emph{this} [object class]?
    \item What action is \emph{this} [object class] doing?
\end{enumerate}
We generate the dataset from Visual Genome~\cite{krishna2017visual}, which consists of 108,249 images with each image annotated on average with 35 objects, 26 object attributes, and 21 pairwise relationships between the objects.

%We consider three question templates: (1) ``What color is this [object class]?'' (2) ``What shape is this [object class]?'' and (3) ``What is this [object class] doing?''

\smallsec{Attribute selection} We select the 100 most common attributes from Visual Genome (e.g. red, round, small), which account for 70.1\% of  annotations. We manually group these attributes into four general categories: color, shape, action, and size. We only use the first three categories, as size is a relative rather than an absolute property of the object.

\smallsec{Question Generation} On every image in Visual Genome, we search for examples that satisfy the constraints of \localqa. Concretely, we examine pairs of bounding boxes $b_i,b_j$ annotated with corresponding object classes $o_i,o_j$ and the sets of attributes $A_i,A_j$. If $o_i=o_j$, so the object classes are the same, we then look for a pair of attributes $a_i\in A_i,a_j \in A_j$ such that %$a_i\in A_i, a_i \notin A_j; a_j \in A_j; a_j \notin A_i$ 
$a_i \neq a_j$ and $a_i,a_j$ are of the same category (color, shape or action). We construct a question $q$ of the form ``What $C$ is this $O$?", where $C$ is the attribute category and $O=o_j=o_k$ the object class. We then add two examples into our dataset: (1) question $q$, point to the center of the bounding box $b_i$, and answer $a_i$, and (2) question $q$, point to center of $b_j$ and answer $a_j$. 

We performed several optimizations to ensure quality. Instances where IoU$(b_i, b_j) \geq t$ were filtered out (we set $t = 0.2$ for high precision). Any object with more than one attribute in the same category was excluded (e.g., a striped shirt annotated as both ``red'' and ``white'') to avoid confusion. Visually similar or identical attributes (e.g., `blonde', `yellow') were collapsed into one attribute to reduce noise.

\begin{figure}[t]
\centering
\includegraphics[width=\linewidth]{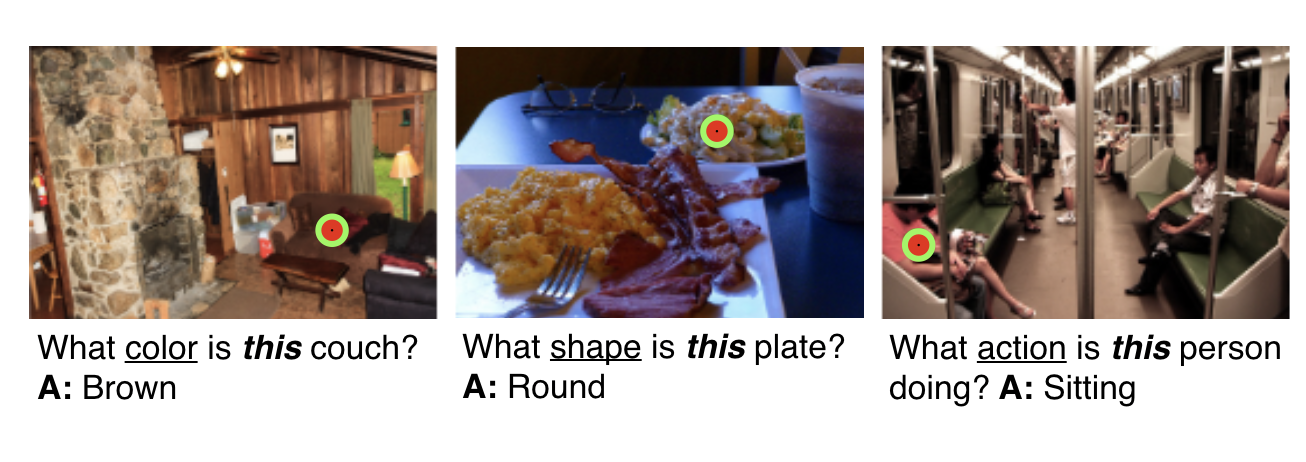}
\caption{Examples of questions in the \localqa Dataset across color, shape, and action. Point is in red.}
\vspace{-1mm}
\label{fig:localqa-data}
\end{figure}

\smallsec{Statistics} The final dataset consists of 57,628 questions across 18,380 images, with 20 unique answers. Due to the high representation of color attributes in Visual Genome, 97.2\% of questions are about color, with the remaining being about shape or action. Although biased in this way, the relative simplicity of color questions allows us to focus on the challenge of incorporating a point input. Although most questions have exactly two unique answers without a disambiguating point, 11.9\% of questions have greater than two answers. Fig.~\ref{fig:localqa-data} shows example questions in the dataset which illustrate the challenges of answering point-input questions, especially when the point can be placed on a non-relevant part of an object (Fig~\ref{fig:localqa-data}a, Fig~\ref{fig:localqa-data}c) or on another smaller, occluding object (Fig~\ref{fig:localqa-data}b). 

The images are randomly divided into (1) \emph{train}, 70\%, with 40,409 questions across 12,867 images; (2) \emph{val}, 10\% with 5,724 questions across 1,838 images; (3) \emph{test-dev}, 10\%, with 5,673 questions across 1,838 images, and (4) \emph{test-final}, 10\%, at 5,910 questions across 1,837 images. 
 % We recommend using \emph{val} for hyperparameter tuning, \emph{test-dev} for ablation studies, and \emph{test-final} to report results. 

\smallsec{Human accuracy} One final question is whether simulating the pointing as simply the center of the bounding box is a sufficiently understandable spatial disambiguation cue, and whether the dataset as a whole is reliable. To evaluate this, we run a small human study on 100 random questions of \emph{test-final}. Please see Appendix \ref{sec:localhuman} for details on human evaluation.

% Humans agreed with the dataset labels $76\%$ of the time, successfully understanding the reference.

\subsection{\localqa models}
\label{sec:localqa-model}

We modify the commonly-used Pythia model ~\cite{Anderson2017up-down,pythia18arxiv} to incorporate the point input $(x,y)$. We consider the more complex transformer-based models in Sec.~\ref{sec:generalqa}; here the language variations of the \localqa dataset are not particularly complex and thus we choose a somewhat simpler and easier-to-analyze model. % We first begin by summarizing the Pythia models.  

\smallsec{Background} Standard VQA systems~\cite{Anderson2017up-down,pythia18arxiv,yu2019mcan} represent the image using a set of bounding box proposals from an object detection model, typically Faster-RCNN~\cite{Ren2015FasterRT}. Concretely: (a) a Region Proposal Network extracts a set of candidate regions along with their ``objectness'' scores, (b) non-max suppression is performed to remove similar boxes, and (c) the top remaining $N$ regions are processed by an \texttt{ROIPool} operation to generate a set of visual features $\mathbf{v}_1\dots \mathbf{v}_N, \mathbf{v}_i \in \mathbb{R}^D$. In step (d), the question is encoded by a recurrent model into $\mathbf{q}\in\mathbb{R}^M$. Then: (e) $\mathbf{q}$ is combined with $\mathbf{v}_1\dots \mathbf{v}_N$ to compute a normalized attention vector $\mathbf{a}\in\mathbb{R}^N$ over the $N$ region proposals, (f) the probabilistic attention $\mathbf{a}$ is used to compute a weighted average of the region features and obtain the image representation $\mathbf{v}\in\mathbb{R}^D$, and (g) $\mathbf{q}$ and $\mathbf{v}$ are projected to a common vector space, combined via element-wise multiplication followed by fully-connected layers and a softmax activation to predict over the answer distribution. Steps (a)-(c) use a pretrained detection model, while (d)-(g) are trained. 
% compute the distribution over the multiple choice answers. 

% The system is trained end-to-end.

\smallsec{Modifications} We choose the simplest way of incorporating the point input into the system above: in step (a) above, we remove all candidate regions that do not contain the point $(x,y)$. The rest of the pipeline is unchanged, apart from zero-padding to $N$ regions if necessary (we use $N = 100$, as is standard). Our approach produces the effect of a pointing gesture by restricting the view of the model to objects around the point. It has the additional advantage of allowing us to analyze the learned attention around the point, yielding insights into the behavior of the model. 

%In Sec.~\ref{sec:localqa-results} we investigate a number of alternatives (e.g., using the $\mathbf{v}_i$ with the highest objectness score of all regions containing the point as the image representation $\mathbf{v}$). 

%we set $\mathbf{v_i}=\mathbf{0}$ for every region $i$ which does not contain the point the point $(x,y)$. 

% \smallsec{Grid-based features} An alternative to region-based features is the newly-proposed grid-based system. Jiang et al.~\cite{jiang2020defense} demonstrated that grid features extracted directly from the convolutional layers of a Faster-RCNN (modified with $1 \times 1$ \texttt{ROIPool} during training) can similarly yield strong performance on VQA; concretely, they replace steps (a-c) above with visual features $\mathbf{v}_{ij}, i\in[1,W],j\in[1, H]$ for downsampled image of resolution $W\times H$. In our grid-based model we incorporate the point input $(x,y)$ by setting the full-image representation $\mathbf{v}=\mathbf{v}_{ij}$ for $ij$ corresponding to the grid position closest to $(x,y)$ after downsampling. %As in the \localmodeln, we empirically evaluate several alternatives in Sec.~\ref{sec:experiments} that learn attention over the grid features contained within some downsampled bounding box (e.g., highest-scoring region proposal). However, we find our method to be most effective.

\subsection{\localqa evaluation}
\label{sec:localqa-results}

We benchmark the model extensions on the \localqa dataset, focusing on insights into the new task.

%We explore several methods for incorporating a point input into the model. Note that for questions in \textbf{PointQA-Local} for which a local region around the point is sufficient, a valid approach is to simply modify existing VQA models to only consider region proposals containing the point. We term this \textbf{LocalQA-Model} compared to our PointQA-Model described in Section~\ref{sec:models} which treats the point as a separate stream of input. We benchmark these and several simpler approaches to incorporating the point. Our analysis is primarily carried out using the SoTA Pythia model, which performs competitively on benchmark VQA datasets (see as well Sec. 6 for comparison). 

\smallsec{Implementation details} We train the model on the \emph{train} subset of the data. For extracting the region features, we use a Feature Pyramid Network (FPN)~\cite{Lin2017FeaturePN} with a ResNet-101~\cite{he2016deep} backbone and IOU of 0.5 for non-maximum suppression, as in the VQA Challenge implementation of Pythia~\cite{pythia18arxiv}. The model was trained using AdaMax with a learning rate of 0.002. We use early stopping on the \emph{val} set with patience of 500 iterations.

% For the grid-based model we compute grid features using the modified Faster-RCNN in \cite{jiang2020defense}. A ResNet-101~\cite{he2016deep} backbone is used for both models. Both models were trained using the AdaMax optimizer with a learning rate of 0.002. We use early stopping on the \emph{val} set with patience of 500 iterations.  

% implementation of \cite{jiang2020defense} where grid features are computed using a modified Faster-RCNN~\cite{Ren2015FasterRT}.

% \localmodel we build off of the implementation of the state-of-the-art Pythia model~\cite{pythia18arxiv}, similarly using a Feature Pyramid Network (FPN)~\cite{Lin2017FeaturePN} to compute region features and IOU of 0.5 for non-maximum suppression. For \gridmodel we use the implementation of \cite{jiang2020defense} where grid features are computed using Faster-RCNN~\cite{Ren2015FasterRT}. For fair comparison, a ResNet-101~\cite{he2016deep} backbone is used for both models. All models were trained using the AdaMax optimizer with a learning rate of 0.002. Early stopping on the \emph{val} set with a patience of 500 iterations.   

\smallsec{Ablation studies} Table~\ref{table:localqa-results} reports results on \emph{test-dev} where we experiment with several different methods for incorporating a point input. Our strategy of removing regions not containing the point and allowing the model to learn attention over the rest achieves at least 5.8\% improvement over simpler strategies such as removing all but the smallest or the highest-scoring region proposal containing the point. Improvement is particularly  significant on action questions, where other methods perform 12.6\% worse.

\begin{table}[t]
\centering
\begin{tabular}{l@{\hskip 0.05in}l@{\hskip 0.03in}c@{\hskip 0.05in}c@{\hskip 0.05in}c@{\hskip 0.03in}c@{\hskip 0.03in}c}
                       & Strategy                  & \multicolumn{1}{l}{QIPBA} & Overall & Color & Action & Shape \\ \hline
\multirow{2}{*}{Priors} & Q-only    & + - - - -                             & 27.8   & 25.8 & 45.5  & 32.1 \\
                    & Modal-A  & - - - - +          
                        & 52.2   & 52.0 & 60.1  & 52.8 \\ \hline
% \multirow{4}{*}{Grid}   & Full Img         & + + - - -                             & 35.7   & 35.4      & 45.5       & 35.9     \\
%                       & Top-score  & + + + - -                             & 50.8   &  51.0     & 45.5  & 35.9 \\
%                         & Smallest         & + + + - -                             & 63.6   & 64.4      & 45.5  & 32.1 \\
%                         & Ours & + + + - -                             & \textbf{64.0}   & 64.4      & 58.0       &  37.7     \\ \hline
                 & Full Img       & + + - - -                             & 37.4   & 37.2      & 45.4       & 35.9      \\
                       & Top-score  & + + + - -                             & 44.8   & 44.4      &  62.2      & 43.4      \\
Model                 & Smallest         & + + + - -                             & 69.2   &  69.6     &  58.0      &  56.6      \\
                        & Ours    & + + + - -                             & \textbf{75.0}   & 75.4 & 66.4  & 56.6 \\  \cline{2-7}
                        & GT box & + + - + -                             & 80.2   & 80.7 & 67.1  & 60.4 \\
                        \hline
\end{tabular}
\caption{Accuracy (in \%) of different models and baselines on the Local-QA \emph{test-dev} split. The QIPBA column indicates the model input: {\bf Q}uestion, {\bf I}mage, {\bf P}oint,  ground truth {\bf B}ounding box, and/or the set of all correct {\bf A}nswers in the image. Modal-A is a baseline oracle that selects the mode answer among all answers that are \emph{correct} in this image. Q-only relies only on dataset language priors;  Full Img is the original VQA model, not using the point input; Top-score and Smallest take the features of only the highest-scoring or smallest region proposal containing the point respectively (thus restricting the attention to a single region). Our proposed strategy outperforms these alternatives and even performs close to when the ground truth box instead of the point is provided during training and testing (bottom row).} 
\label{table:localqa-results}
\end{table}

\begin{figure}[t]
\centering
   \includegraphics[width=1\linewidth]{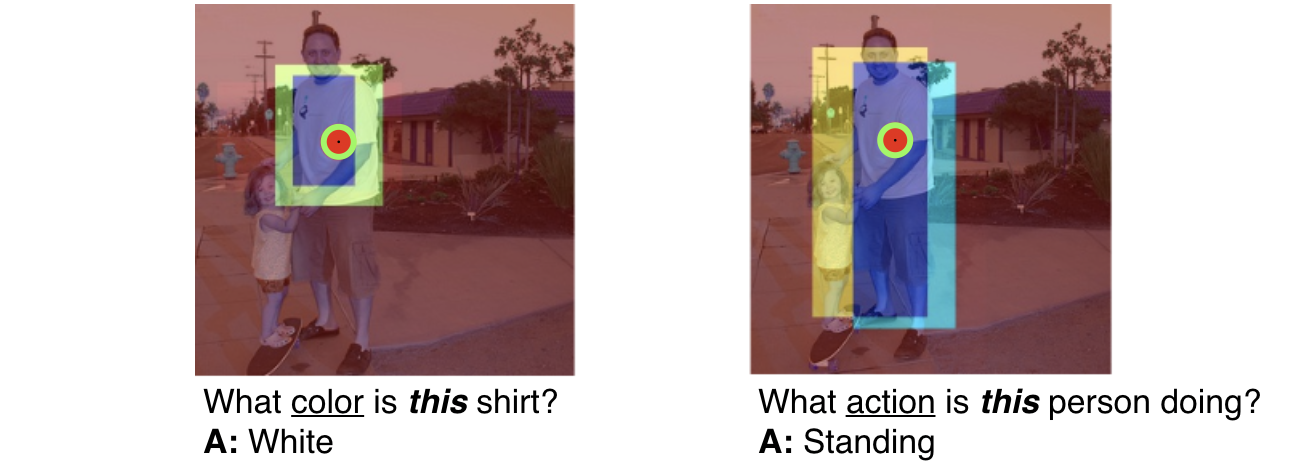}
   \caption{The Pythia-based  local region model (Sec.~\ref{sec:localqa-model}) successfully varies attention around the point in response to a question. (Darker = higher attention; point in red).}
\label{fig:localqa-attention}
\end{figure}

% We analyze the attention of the model around the point. 
\smallsec{Attention analysis} An average of 27 region proposals containing the point are provided as input to the model. The attention is relatively peaked, with the maximum attention of 0.548 on average in \emph{test-dev}.  An important property is the model's ability to attend based on the question. For the 799 questions in \emph{test-dev} that ask ``What color is this shirt?", we can change the question to ``What action is this person doing?'' This increases the median size of the max-attention region from 2,997 to 5,451 pixels, indicating that the model shifts its attention accordingly. (Fig.~\ref{fig:localqa-attention}).  % update

%\textcolor{blue}{Would it be good to re-run this analysis for the PointQA-Model?}

% The size of the maximum attention region is only 71\% the size of the ground-truth object bounding box on average, likely because smaller bounding boxes are sufficient to infer attributes such as color and action.

\begin{figure}[t]
\centering
   \includegraphics[width=\linewidth]{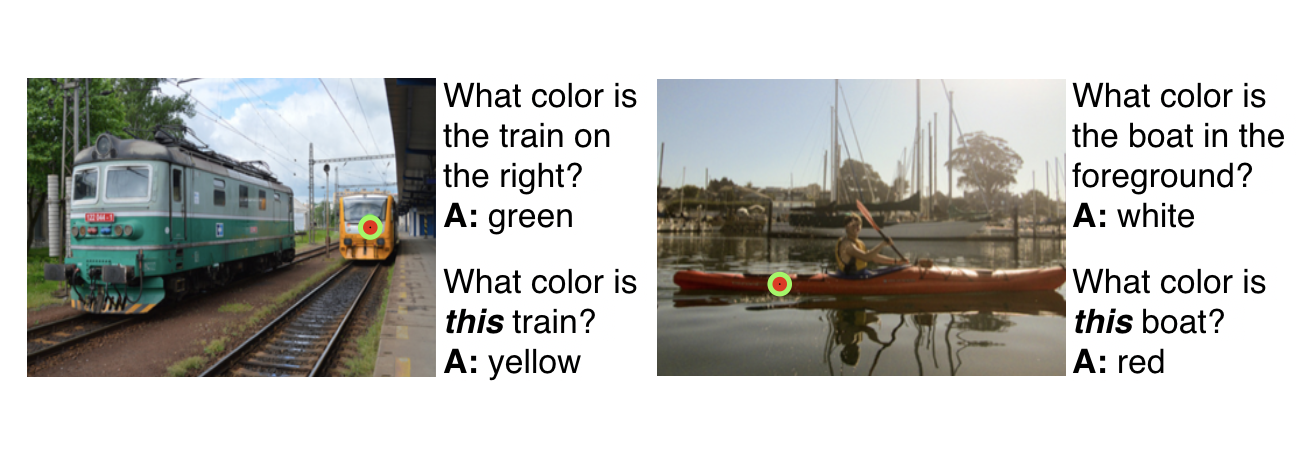}
   \caption{In both examples, the standard verbal-only VQA model (top) fails to understand the verbal disambiguation and picks an answer corresponding to the wrong object instance; ours (bottom) correctly incorporates the point input.}
\label{fig:spatial-vs-verbal}
\end{figure}

\smallsec{Spatial vs Verbal Disambiguation} Finally, we note for many \localqa questions, a disambiguating phrase could be provided as a substitute for a point (e.g. ``What color is the car \emph{on the left}?'') Such verbal disambiguation is often unnatural and can grow expensive depending on the object of interest; this partially explains why humans prefer to point and motivates pointing as a more realistic setup for a broad set of visual questions. We provide a brief comparison between verbal and spatial disambiguation.

\emph{Data.} To do so, we collect a small dataset using human-written questions from Visual Genome~\cite{krishna2017visual}. A parser is used to detect the question subject, and we (1) ensure this object appears multiple times in the image and (2) search for prepositional phrases appended to the question subject to detect verbal disambiguation. Using object/attribute annotations, the question's subject is matched to an object in the image, and a ``spatial version" of the question is generated by removing the verbal disambiguation  and instead generating a point from the center of the target object's bounding box. We thus obtain a verbally disambiguating version of the dataset $D_V$ and a spatial disambiguating version $D_S$, with 1,855 questions across 1,575 images in each. We use 80\% for training and 10\% each for validation and testing. %Further details of dataset construction can be found in the Appendix.

\emph{Results.} Our model leveraging the point input achieves an accuracy of 66.5\% on $D_S$, while the Pythia~\cite{pythia18arxiv} baseline relying on verbal disambiguation achieves only 26.5\% on $D_V$ (Fig.~\ref{fig:spatial-vs-verbal}). Note that a question-only model gets 22.2\% accuracy on $D_V$; thus, the baseline does not effectively understand verbal disambiguation. This further motivates point-input questions as an extension of VQA.

% Notably, training Pythia using the questions in $D_S$ while providing the entire image achieves 32.28\% accuracy, higher than the verbal model. This indicates that the model is effectively unable to understand verbal disambiguation, and that it simply adds noise to the question that reduces the model's accuracy. Fig. 6 shows an example where the verbal model is incorrect. On top of this accuracy gap, the practical value and fundamental challenges of point-input questions motivate further investigation of this question space. 

\section{PointQA-LookTwice: reasoning about a local region in the broader image context}
\label{sec:looktwiceqa}

We next consider the more general \looktwiceqa setting which requires situating a local region in the broader context of the image. A natural example is counting questions such as ``How many of \emph{this} animal are there?", where the model must identify the relevant object around the point and then use this information to attend to the entire image. Correspondingly, in Section \ref{sec:looktwiceqa-dataset} we construct a dataset of counting questions from Visual Genome \cite{krishna2017visual} such that looking at a local region around the point is insufficient to answer the question. We then introduce a new model in Section \ref{sec:looktwice-models} that includes global attention and show that it successfully reasons about the point in the broader image context.

\subsection{\looktwiceqa dataset}
\label{sec:looktwiceqa-dataset}

%. We augment these using questions automatically generated from Visual Genome object annotations; however, these are prone to under-counting when the object annotation recall is low and are thus only utilized during training.

%where the model needs to first understand the local region around the point and then use this local information to reason about the rest of the image. A natural example is counting questions. A standard VQA counting question (e.g. ``How many trucks are there?") can be rephrased to include a point input as (1) ``How many of \emph{these} \underline{vehicles} are there?", or more generally (2) ``How many of \emph{these} are there?". Such questions require both extracting appropriate local features as well as reasoning globally to count the instances.

%and explore the robustness of our point-based VQA model to increasing generality of the question.

%As the question becomes increasingly general, the point input .

We construct the \looktwiceqa dataset leveraging  the 99,860 human-written counting questions in Visual Genome~\cite{krishna2017visual} (e.g., ``how many trucks are there?''). We turn it into pointing QA by further leveraging the object annotations as in Sec.~\ref{sec:localqa-data}. 

\smallsec{Question Generation} For each question, we extract the subject of the question (e.g., ``truck'') and attempt to match it to an annotated region in the image. If multiple regions exist, one is chosen at random; if none exist, the question is removed. %For all questions, the point is taken as the center of this object's bounding box.
From this filtered question set, questions about object classes appearing less than 100 times total are further removed. The remaining set of object classes are manually grouped into three super-categories: beings (people and animals), vehicles (ex. cars, planes), and objects (ex. laptops, umbrellas).
From each question of the form e.g., ``How many  [object class] are there?'' we generate two new questions by replacing the ``[object class]'' either with (1)  ``these [supercategory]'' or (2) just ``these.'' We disambiguate the reference through the point input, simulated as the center of the object bounding box. The answer remains the same for all three questions; we thus train the model to locally infer the object category from the point and then count globally. From the raw object counts we bin the possible answers to ``1", ``2", and ``$>$ 2", such that the model predicts one of three answers to the counting question. 

\smallsec{Counteracting priors} Our goal is to construct questions that \emph{require} the point input and cannot be answered without it. Thus, we enforce that for each image $I$ in the evaluation set, there are at least two questions $q_i, q_j$ such that $o_i \neq o_j$ (different objects), and $a_i \neq a_j$ (different answers). 

The remaining generated questions are added to the training set. To prevent priors in the training set, for each question $q_i$ we use Visual Genome object annotations to generate a question $q_j$ with $o_i \neq o_j$ and $a_i \neq a_j$. Object counts are generated from the IoU-filtered object annotations in the image and may be imperfect due to under-counting (thus not used during evaluation). The original question is constructed as ``How many $o_i$ are there?", with the object class in its plural form for readability, and then converted into two generic question types as above.

\smallsec{Statistics} The \looktwiceqa dataset contains 57,405 questions across 34,676 images, including (1) \emph{train} with $37,981$ human-written and $14,713$ automatically-generated questions across 32,925 images, (2) \emph{val} with $997$ human-written questions across 380 images, and (3) \emph{test} with $3,714$ human-written questions across $1,371$ images. 

%Specifically, the evaluation set, where we explicitly require questions with different answers for each image, contains 4,711 questions. 1,000 questions are reserved for the validation set, and the rest are included in the final test set. Within the training set, 14,713 questions are derived from Visual Genome annotations. 

%Note that while we derive questions from annotations for training, the evaluation set consists exclusively of human-written questions.

\begin{figure}[t]
\centering
\includegraphics[width=\linewidth]{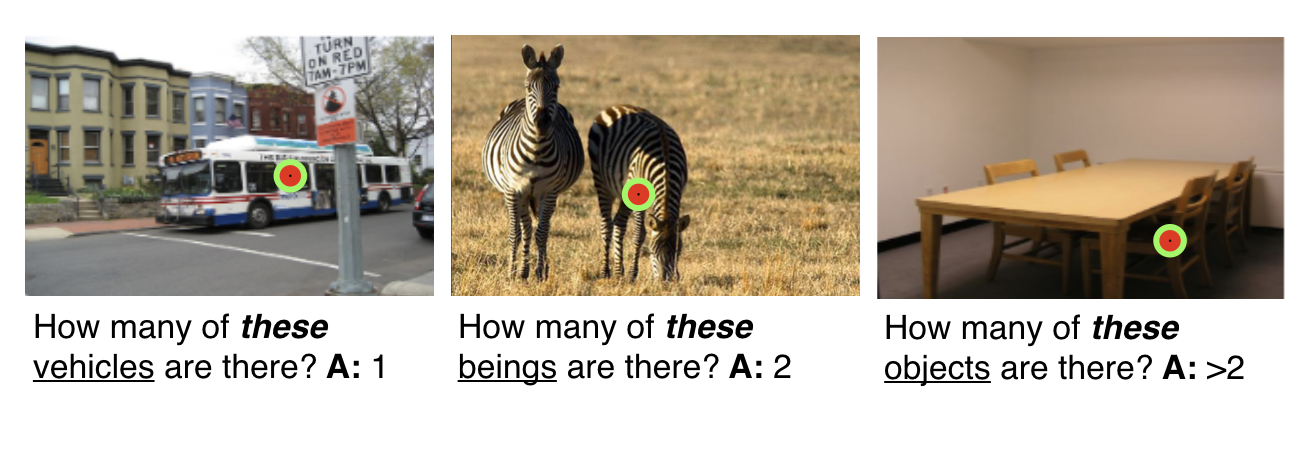}
\caption{Example entries in the \looktwiceqa Dataset. Point is in red and supercategory is underlined.}
\label{fig:looktwice-data}
\end{figure}

The answer distribution is reasonably balanced: the answers ``1'', ``2'', and ``$>$ 2'' appear with $35.4\%$, $33.7\%$, and $30.9\%$ frequency respectively in the test set. The most common objects in the dataset are people ($29.7\%$), cars ($3.1\%$), and signs ($2.9\%$). Due to the high representation of people, the most common super-category is beings ($48.9\%$), followed by objects ($36.7\%$) and vehicles ($14.3\%$). Given only the object name, a model can achieve an accuracy of at most $42.1\%$, indicating that visual reasoning is necessary. Example questions are shown in Fig.~\ref{fig:looktwice-data}.

\smallsec{Human accuracy} As in Sec.~\ref{sec:localqa-data} we run a small-scale human study on 100 \emph{test} questions to evaluate the reliability of the data. Please see Appendix \ref{sec:lthuman} for details on human evaluation. %, suggesting that the counting task with a point is somewhat more difficult/ambiguous but still highly doable. 

% Humans agreed with the answer $79\%$ of the time, suggesting that the counting task with a point is doable. $

% When the answers differ, on average humans count 1.3 more objects, suggesting that the annotations used for counting may be somewhat incomplete. 

\subsection{\looktwiceqa model}
\label{sec:looktwice-models}

\smallsec{Local-only} We adapt the Pythia-based model of Sec.~\ref{sec:localqa-model} to this new setting, producing a model that combines local with global contextual reasoning. So far we had a set of visual features $\{\mathbf{v}_i^{pt}\}$ corresponding to region proposals containing the point $pt$, and we computed an attention vector over these proposals. Concretely, we projected the visual features $\{\mathbf{v}_i^{pt}\}$ and the text features of the question $\mathbf{q}$ to a common space, multiplied them element-wise and then ran through fully-connected layers to produce an attention vector, which is then normalized to a probability distribution $\mathbf{a}^{pt}$. This allowed us to produce $\mathbf{v}^{pt} = \sum_i \mathbf{a}_i^{pt}\mathbf{v}_i^{pt}$ describing the relevant information of the local region. 

\smallsec{Global attention} Now we further use this local information to attend to the entire image: concretely, we project the visual features $\{\mathbf{v}_i\}$ from \emph{all} region proposals, the text features $\mathbf{q}$, and the new $\mathbf{v}^{pt}$ to a common subspace, and then as before element-wise multiply and run through fully-connected layers to compute the normalized attention $\mathbf{a}^{all}$, yielding the global representation  $\mathbf{v}^{all} = \sum_i \mathbf{a}^{all}_i\mathbf{v}_i$. The intuition is that $\mathbf{v}^{pt}$ extracts local information relevant to the question, which is then used to attend to relevant regions in the whole image. Given these three streams of information, $\mathbf{q}$, $\mathbf{v}^{pt}$ and $\mathbf{v}^{all}$, we combine them through pairwise element-wise multiplication and concatenation (multiply $\mathbf{q}$ and $\mathbf{v}^{pt}$, and $\mathbf{q}$ and $\mathbf{v}^{img}$, and concatenate the two results) followed by fully connected layers and a softmax output (step (g) in Sec.~\ref{sec:localqa-model}).

\subsection{\looktwiceqa evaluation}
\label{sec:looktwice-results}

We benchmark the new global model trained on the \emph{train} set of \looktwiceqa using the setup from Sec.~\ref{sec:localqa-results}. 

\smallsec{Test accuracy} When trained to answer the question ``How many of \emph{these} are there?'' along with a disambiguating point input, the model achieves an accuracy of 56.5\%, significantly higher than an image-only version (without the point) at 46.1\% and the modal answer (``1'') baseline at 35.4\%. As expected it is somewhat behind having access to the object's bounding box at 60.2\%. In Table~\ref{table:globalqa-accuracy} we demonstrate how the accuracy changes with decreasing the ambiguity of the question: specifying the supercategory (being, vehicle or object) boosts the accuracy from 56.5\% to 59.1\%, and naming the object class further boosts it to 62.8\% (in fact, as expected making the need for spatial supervision irrelevant, as the image-only model achieves 62.7\% in this setting).

%providing the object class in the question increases the accuracy of \looktwicemodel only from $51.1\%$ with a supercategory to  

%We primarily 

%Results are shown in Table 3. When increasing the ambiguity of the object reference in the counting question (object $\rightarrow$ super-category $\rightarrow$ no object), the accuracy of Pythia without spatial disambiguation drops from 52.57\% eventually to 39.61\%. Providing the bounding box of the referred object along with the question recovers this drop (only 52.41\% to 52.30\%), proving robust to increasing question ambiguity as expected. Notably, our Global-QA model, which accepts a point input, is also robust to increasing question ambiguity. When asking ``How many of these are there?", our model achieves 49.62\% accuracy, significantly higher than the model with no spatial disambiguation and only a 2.57\% drop in accuracy from when the object name is provided.

\begin{table}[t]
\centering
\begin{tabular}{lccc|c}
& \multicolumn{3}{c|}{Spatial Disambiguation}&\multirow{2}{*}{Prior}
\\
& None       & {\bf Point}      & Box &    \\ 
 \hline
How many of \emph{these}...  & 46.1 & 56.5 & 60.2 & 37.8    \\
\begin{tabular}[c]{@{}l@{}}How many of \emph{these}\\ {[}supercategory{]}...\end{tabular}  & 53.1       &  59.l    & 59.8 & 38.6          \\
How many {[}object{]}...  & 62.7      & 62.8     & 61.4 & 40.3          \\
\hline
\end{tabular}
\caption{Results of the Pythia-based global model (Sec.~\ref{sec:looktwice-models}) on the  \looktwiceqa \emph{test}. Rows are the question asked; columns are the disambiguation provided. Prior is a language-only model.}
\label{table:globalqa-accuracy}
\end{table}

\begin{figure}[t]
\centering
   \includegraphics[width=\linewidth]{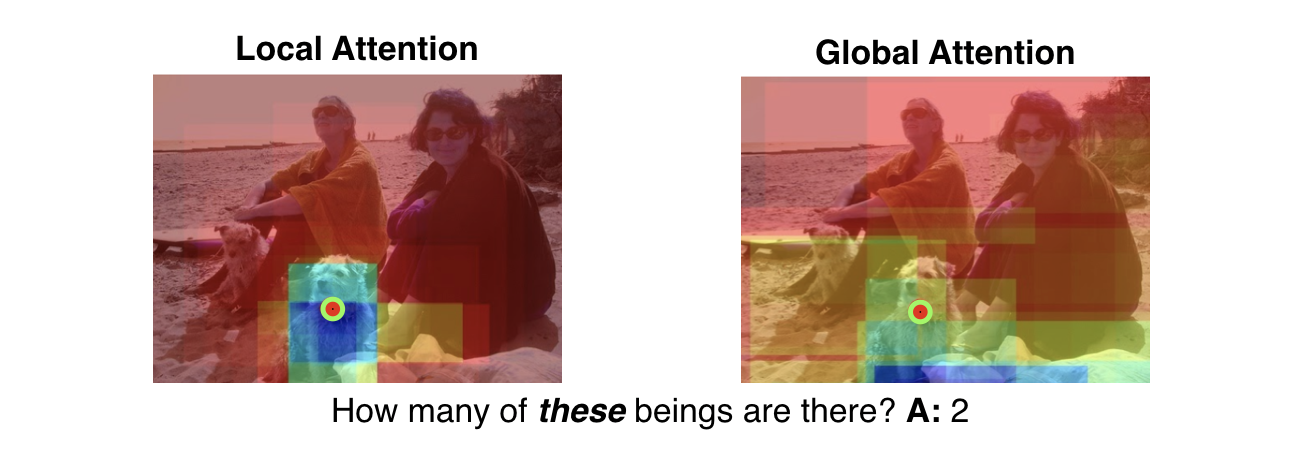}
   \caption{Attention of the Pythia-based model on a question in the \looktwiceqa \emph{test} set. Point is in red.}
\label{fig:lt_qual}
\end{figure}

\smallsec{Attention analysis} We examine the model trained on supercategory questions. There are two types of attention: attention around the point is relatively peaked, with an average max attention of 0.42; as expected, global attention on the image is diffuse with an average max attention of 0.07. An example result is shown in Fig. \ref{fig:lt_qual}.

\smallsec{Global vs local-only attention} We demonstrate the need for the new global attention added in Sec.~\ref{sec:looktwice-models} by comparing with the previous model which only considers the local regions around the point. The local-only model achieves an accuracy of 55.9\% on supercategory questions, significantly lower than the global model at 59.1\%. As expected, this is primarily due to under-counting (since the local-only model is not attending to other object instances). 
Concretely, on questions with the answer ``1'', the global model performs slightly better than the local-only model (75.64\% vs. 74.12\%); however the difference is greater for answer $\geq$2, where the global model achieves 50.04\% and local model only 45.84\%. Finally, as a sanity-check, we verified that the models achieve equal accuracy on the simpler \localqa dataset, indicating that the global model is well-suited for a wider range of pointing-based questions.

\section{PointQA-General: generalized reasoning from a point input}
\label{sec:generalqa}

Equipped with the insights of Sec.~\ref{sec:localqa} and \ref{sec:looktwiceqa}, where we examined well-structured pointing questions, we now turn our attention to the unconstrained setting where the questions become much more complex and the models must reason about image, textual, and spatial input in full generality. We construct a new dataset, modify state-of-the-art transformer-based models, and perform analysis in this new setting to round out our exploration. 

\subsection{\generalqa dataset}
We generate the \generalqa dataset by adapting the human-written questions from Visual7W~\cite{zhu2016visual7w}.

%: these questions are paired with four bounding boxes, one of which is correct. 

%\generalqa dataset 

%V7W-Point consists of questions reformulated from Visual7W’s pointing questions, on the premise that pointing questions require disambiguation in the form of a point to refer to the specific object instance in question.

\begin{figure}[t]
\includegraphics[width=\linewidth]{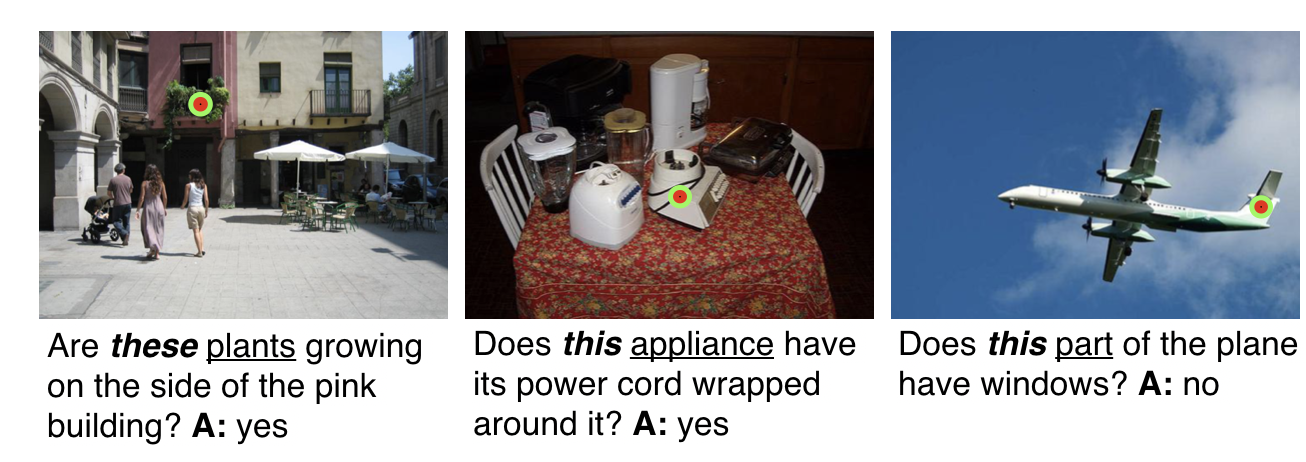}
\caption{Examples entries in the \generalqa Dataset. Point is indicated in red.}
\label{fig:generalqaex}
\end{figure}

\smallsec{Question Generation} ``Which'' questions in Visual7W are paired with bounding boxes as multiple choice answers; for example, they might ask ``Which pillow is closest to the window?'', where one of the four provided boxes is correct.  We transform these questions into pointing-QA using the following formula (where $X$ is the subject and $Y$ is a description):
\begin{itemize}
    \item ``Which $X$ is $Y$'' becomes ``Is this $X$ $Y$?'' 
    \item ``Which $X$ are $Y$'' becomes ``Are these $X$ $Y$?'' 
    \item ``Which $X$ has $Y$'' becomes ``Does this $X$ have $Y$?'' 
    \item ``Which $X$ have $Y$'' becomes ``Do these $X$ have $Y$?''
\end{itemize}
A high proportion of human-written questions follow one of these templates; questions that do not are not included. From each included question in Visual7W we generate two pointing questions: (1) using the correct bounding box, with the answer ``yes'' and (2) using one of the three incorrect boxes selected randomly with the answer ``no''. The point input is simulated as the center of the bounding box. Example questions are shown in Fig. \ref{fig:generalqaex}.

%Of the available  questions, we take the subset that are of the form: "Which" + \{...\} + “is/are/has/have” + \{...?\}. We label these four parts $a_i$, $b_i$, $c_i$, $d_i$, respectively, for Visual7W question $Q_i$. We transform $Q_i$ into V7W-Point question $P_i$ using the following formulation: If $b_i$ is "is" or "are," then $P_i = c_i$ + "this/these" + $b_i$ + $d_i$. If $b_i$ is "has" or "have," then $P_i = $ "Does this/Do these" + $b_i$ + "have" + $d_i$. For each $P_i$, one of the four bounding boxes paired with $Q_i$ is randomly chosen with 0.5 probability of choosing the correct box. The answer for each $P_i$ is either "yes" or "no," depending on which box is chosen.

\smallsec{Statistics} The final dataset consists of 319,300 questions over 25,420 images, with the ``yes'' and ``no'' answers equally balanced. The dataset is divided into (1) \emph{train} (255,072 questions across 20,337 images), (2) \emph{val} (32,276 questions across 2,541 images), and (3) \emph{test} (31,952 questions across 2,542 images). The most common subjects asked about are ``part'' (e.g. ``Is \emph{this} part of the desktop computer touching the carpet?''; 6.2\% of the questions), ``object'' (4.4\%), ``person'' (4.3\%),  ``item'' (2.7\%), ``man'' (1.6\%), ``tree'' (1.3\%), ``animal'' (1.2\%), and ``food'' (1.0\%). The reference descriptions are quite complex: they contain 6 words on average ($95\%$ fall between 3 and 12 words) compared to the much simpler attributes of \localqa which are typically only 1 word long.

\smallsec{Human accuracy} As for the previous two datasets we ran a small-scale human study on 100 \emph{test} questions. Please see Appendix \ref{sec:generalhuman} for details on human evaluation.

% Human accuracy on 100 random \emph{test} questions is 91\%; this confirms both our assumptions about simulating the point as the center of the bounding box as well as the reliability of the overall data collection pipeline. 

\subsection{\generalqa models}
\label{sec:generalqa-model}

% \begin{figure}[!t]

% \centering
% \includegraphics[width=0.06\textwidth,height=.15\textwidth]{latex/quesimg.png}\hfill
% \includegraphics[width=.16\textwidth,height=0.15\textwidth]{latex/quesimgpoint.png}\hfill
% \includegraphics[width=.17\textwidth]{latex/threewayatt.png}
% \caption{Insert caption here.}
% \label{fig:figure3}
% \end{figure}

\begin{figure}[!t]
\centering
\includegraphics[width=.065\textwidth,height=.16\textwidth]{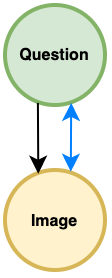} \hspace{5mm}
\includegraphics[width=.16\textwidth]{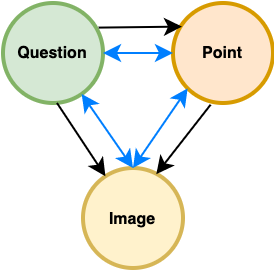}
\includegraphics[width=.15\textwidth]{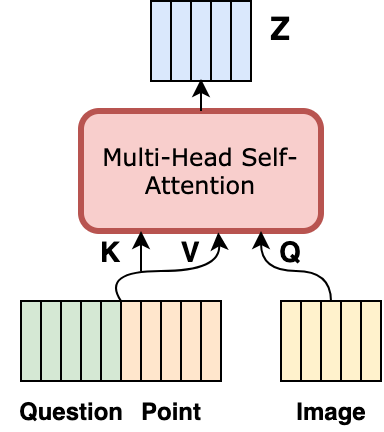}

\caption{Left shows the different attention dependency structures between image and text; middle shows when the ``point stream" is added (Sec. \ref{sec:generalqa-model}). Blue arrows indicate the bidirectional structure and black arrows unidirectional. Right shows the exact way in which cross-attention is implemented for transformer models with three streams.}
\label{fig:dependency}
\end{figure}

Given the increased complexity of the questions in \generalqa we adapt two recent transformer-based VQA models to our pointing-based VQA task, MCAN \cite{yu2019mcan} and LXMERT \cite{tan2019lxmert} (in addition to the Pythia-based model of Sec.~\ref{sec:looktwice-models}). At a high-level MCAN computes a unidirectional attention like Pythia where the text influences attention over the image; by contrast in the bidirectional approach of LXMERT~\cite{tan2019lxmert} each modality influences attention over the other (Fig. \ref{fig:dependency}a). We now describe the model details.

% \begin{figure}[!t]
% \begin{subfigure}{0.2\linewidth}
% %\centering  % redundant
% \includegraphics[width=0.65\textwidth,height=1.8\textwidth]{latex/quesimg.png}

% \caption{Unidir.}
% \end{subfigure}%
% \hfill% not: "\hspace{0.5cm}"
% \begin{subfigure}{0.3\linewidth}
% %\centering  % redundant
% \includegraphics[width=1.07\textwidth,height=1.15\textwidth]{latex/quesimgpoint.png}
% \caption{Bidir.}
% \label{fig:figure2}
% \end{subfigure}%
% \hfill% not: "\hspace{0.5cm}"
% \begin{subfigure}{0.4\linewidth}
% %\centering  % redundant
% \includegraphics[width=0.75\textwidth,height=0.9\textwidth]{latex/threewayatt.png}
% \caption{Cross-attention}
% \label{fig:figure3}
% \end{subfigure}
% \caption{8a shows the different attention dependency structures between image and text; 8b shows when the ``point stream" is added (Sec. \ref{sec:generalqa-model}). Blue arrows indicate the bidirectional structure and black arrows unidirectional. 8c shows the exact way in which cross-attention is implemented for transformer models with three streams.}
% \end{figure}
 
\smallsec{MCAN-based}  MCAN~\cite{yu2019mcan} is the winner of the the 2019 VQA Challenge. It works as follows: (1) The question $\mathbf{q}$ is broken into tokens $\{q_j\}$ and is encoded by stacking several self-attention layers $\{q_{j}\}_{\ell + 1} = A(Q_\ell, K_\ell, V_\ell)$, where $A$ is the self-attention mechanism and $(Q_\ell, K_\ell, V_\ell)$ are the queries, keys, and values derived from $\{q_{j}\}_{\ell}$ at layer $\ell$. (2) The image representation over visual features derived from region proposals $\{\mathbf{v}_j\}$ is computed using a stack of self-attention and cross-attention layers. In each cross-attention layer, the keys and values are derived from the question representation at final layer $L$; in other words, $\{\mathbf{v}_j\}_{\ell + 1} = A(Q_\ell, W_K\{q_j\}_L, W_V\{q_j\}_L)$ for weights $W_K$ and  $W_V$ corresponding to keys and values respectively. (3) The feature representations at the final layer $\{q_j\}_L$ and $\{\mathbf{v}_j\}_L$ are pooled, fused, and fed into a classifier. 
%Refer to \cite{yu2019mcan} for details.

To adapt this model to further accept the point input, we compute visual features $\{\mathbf{v}^{pt}_j\}_L$ corresponding only to the regions containing the point $pt$  (the ``point stream''), conditioned on the question as in the original model. Then in each cross-attention layer for the \emph{image} stream, we condition attention on information from both the question and point by concatenating the representations $\{f_j\}_\ell = \{q_j\}_\ell \oplus \{\mathbf{v}^{pt}_j\}_\ell$ and then computing the keys and values (Fig. \ref{fig:dependency}c). We then compute the multimodal fusion function:

\vspace{-3mm}

\begin{align*}
    &z_1 = \text{LayerNorm}(W_1^T\{q_j\}_L + W_2^T\{\mathbf{v}^{pt}_j\}_L), \\
    &z_2 = \text{LayerNorm}(W_1^T\{q_j\}_L + W_3^T\{\mathbf{v}_j\}_L), \\
    &z = z_1 \oplus z_2,
\end{align*}
followed by a linear classifier over the answer space. This dependency between streams is shown in Fig. \ref{fig:dependency}b (black).

% To adapt this model to further accept the point input, we additionally compute visual features $\{\mathbf{v}^{pt}_j\}_L$ corresponding only to the regions containing the point $pt$  (the ``point stream''). Each cross-attention layer for the point stream is conditioned on the question, while for the visual stream we condition attention on information from both the question and point by concatenating the representations $\{f_j\}_\ell = \{q_j\}_\ell \oplus \{\mathbf{v}^{pt}_j\}_\ell$ and then computing the keys and values. Finally we compute the  multimodal fusion function:
% \begin{align*}
%     &z_1 = \text{LayerNorm}(W_1^T\{q_j\}_L + W_2^T\{\mathbf{v}^{pt}_j\}_L), \\
%     &z_2 = \text{LayerNorm}(W_1^T\{q_j\}_L + W_3^T\{\mathbf{v}_j\}_L), \\
%     &z = z_1 \oplus z_2,
% \end{align*}
% followed by a linear classifier over the answer space.

\smallsec{LXMERT-based} LXMERT implements a bidirectional approach to attention; first, the question and visual features are passed through a stack of self-attention layers to obtain $\{q_j\}_{L}$ and $\{\mathbf{v}_j\}_L$ as above. Then both are passed through a stack of ``cross-modality encoders'' that include a cross-attention layer followed by a self-attention layer and small feed-forward layer. The cross-attention layer exchanges the keys and values of the other modality (exactly as MCAN but for \emph{both} image and text).

% MCAN implements a \emph{unidirectional} approach where the text influences the attention over the image; 
% by contrast in the \emph{bidirectional} approach of LXMERT~\cite{tan2019lxmert} every modality influences attention over the other (Fig. 2a). 

We modify LXMERT similarly to MCAN to accept the point input: it now has three streams corresponding to the question $\mathbf{q}$, the global image stream corresponding to all image regions $\mathbf{v}$, and a local point stream corresponding to only the regions $\mathbf{v}^{pt}$. Concretely, for the cross-attention layer at layer $\ell$ operating on the visual features $\{\mathbf{v}_j\}_\ell$, we concatenate $\{f_j\}_\ell = \{q_j\}_\ell \oplus \{\mathbf{v}^{pt}_j\}_\ell$ and compute keys and values on top of this representation; we do this similarly for the other modalities at layer $\ell$. Thus, at each layer each modality influences attention over the other in the cross-attention modules (Fig. \ref{fig:dependency}b, blue). As in LXMERT the final language features are pooled and fed into a classifier.

\smallsec{Two-Stream vs. Three-Stream} An alternative way of incorporating both local and contextual information is to concatenate the image features $\{\mathbf{v}_j\}$ and the point features $\{\mathbf{v}^{pt}_j\}$ into the same stream without modifying the VQA model itself. We explore this alternative in Sec \ref{sec:generalqa-results} which we term \textbf{two-stream}, compared to the \textbf{three-stream} approach we describe in this section that treats the point as its own stream and modifies the cross-attention accordingly.

\subsection{\generalqa evaluation}
\label{sec:generalqa-results}

We evaluate the Pythia-based~\cite{pythia18arxiv} model of Sec.~\ref{sec:looktwice-models} and the MCAN-based~\cite{yu2019mcan} and LXMERT-based~\cite{tan2019lxmert} models of Sec.~\ref{sec:generalqa-model} on the challenging task of \generalqa. 

\smallsec{Implementation Details}. For MCAN we set $L = 2$, since we found that increasing number of layers did not improve performance. For LXMERT following the original paper, we set a higher number of layers for the language modality; concretely we set $N_L = 5$ and $N_{Img} = N_{Pt} = 3$. We set the number of cross-modality encoders $N_X$ to be 3. Both models were trained using Adam with early stopping and a patience of 5000 iterations. We used a learning rate of $\{5, 2.5\}e{\text -}5$ for MCAN and LXMERT respectively with a warmup/decay schedule. Details for Pythia are in Sec. \ref{sec:localqa-results}. 

\smallsec{Test Results} Results on the \generalqa dataset are shown in Table 3. The single highest accuracy is achieved using our three-stream approach with the MCAN-based model. Across all models, the three-stream approach outperforms any of the ablations that use one (\emph{Q-Only}) or two streams (\emph{Image+Q}, \emph{Point+Q}). The high overall accuracy of our three-stream approach and its improvement over \emph{Point+Q} indicates the benefit of adding the point as a separate stream and modifying the model cross-attention to create a rich set of contextual interactions between the streams. The \emph{Image+Q} and \emph{Q-Only} models perform no better than random chance since for each image-question pair in Visual7W we generate two questions with opposite answers.

\begin{figure}[t]
\centering
\includegraphics[width=0.9\linewidth]{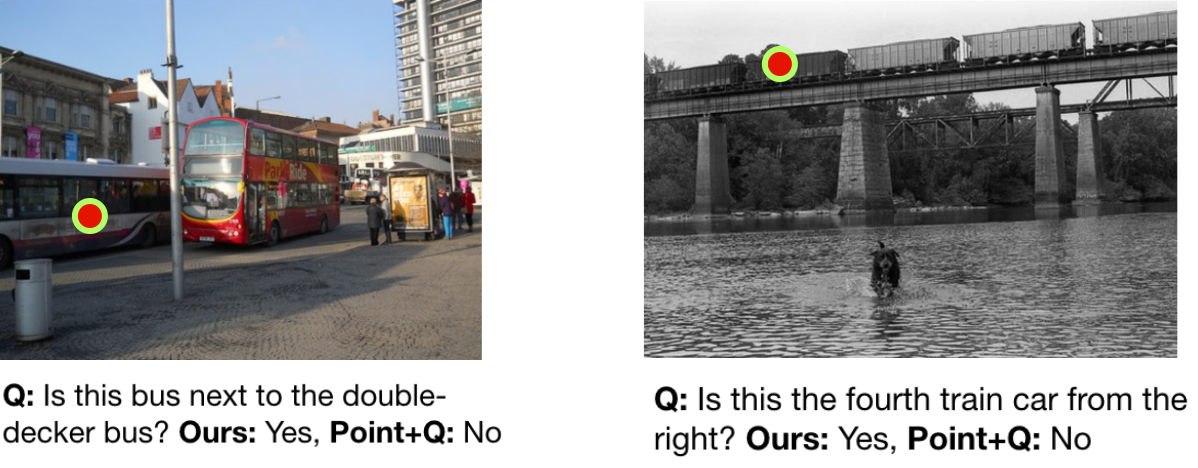}
\caption{Examples in the \generalqa test-set where our three-stream approach (w/ Pythia) is correct and the Point+Q ablation does not have contextual information.}
\label{fig:qualexgeneral}
\end{figure}

The MCAN-based method appears to achieve the strongest performance, even for the \emph{Point+Q} ablation; this suggests it might make more effective use of the limited contextual information available in this setting. For both MCAN and Pythia, the two-stream approach makes less effective use of contextual information than three-stream. Surprisingly, the LXMERT model achieves higher accuracy on the two-stream approach. One possible reason is that the pooling strategy used by LXMERT only uses the language representation; it is possible that pooling from all three streams would improve performance of the three-stream method. Qualitative results are shown in Fig. \ref{fig:qualexgeneral}.

\begin{table}[]
\centering
% \begin{tabular}{llll}
%               & Pythia         & LXMERT & MCAN           \\ \hline
% Q-Only & 50.00          & 50.00  & 50.00          \\
% Image+Q    & 50.00          & 50.00  & 50.00          \\
% Point+Q    & 81.81          & 81.33  & 82.60          \\ \hline
% Two-Stream    & 77.84          & \textbf{82.41}  & 81.62         \\
% Three-Stream  & \textbf{83.12} & 81.71  & \textbf{83.21}
% \end{tabular}

\begin{tabular}{llll}
              & Pythia         & MCAN & LXMERT           \\ \hline
Q-Only & 50.00          & 50.00  & 50.00          \\
Image+Q    & 50.00          & 50.00  & 50.00          \\
Point+Q    & 81.81          & 82.60  & 81.33          \\ \hline
Two-Stream    & 77.84          & 81.62  & \textbf{82.41}       \\
Three-Stream  & \textbf{83.12} & \textbf{83.21}  & 81.71
\end{tabular}

\vspace{3mm}

\caption{\textbf{Accuracy on the \generalqa dataset}. We evaluate the following methods and baselines: (1) Three-Stream and Two-Stream as described in Sec \ref{sec:generalqa-model}, (2) \emph{Q-only} relies only on the language of the question; (3) \emph{Image+Q} provides the question and visual features $\{\mathbf{v}_j\}$ corresponding to the entire image, and (4) \emph{Point+Q} provides the question and only the features $\{\mathbf{v}_j^{pt}\}$ for the region proposals containing the point (for Pythia this is the local-only model of Sec.~\ref{sec:localqa}). Note that all these baselines are outperformed by our proposed models which capture local and global contextual information, and that the highest accuracy is achieved with our three-stream approach on the MCAN model.}
\end{table}

\smallsec{Using Image Context} Certain questions in the \generalqa Dataset require reasoning about the image context beyond a local region around the point, such as questions involving comparison (e.g. ``Is this zebra the \emph{most} obstructed from view?") or directional reasoning (``Are these crayons on a table \emph{near} a cat?"). Viewing only the object being directly pointed to is insufficient to answer these questions. The accuracy improvement of the three-stream model over the \emph{Point+Q} ablation is particularly significant for such questions containing comparison/directional words; e.g. for questions with the word `farthest', the accuracy difference of the two models (w/ Pythia) is 3.66\% vs. 1.81\% overall. This result holds strongly across several such words, suggesting that the improved performance of the three-stream model is due to its ability to effectively incorporate the broader image context. Examples in Fig. \ref{fig:qualexgeneral} also indicate this effect.

\section{Comparison Across Three Settings}
\label{sec:task-comparison}

Having examined three settings, we take the opportunity to briefly articulate two key features that emerge consistently in the PointQA-task: (1) the need for visual grounding, and (2) the challenge of reasoning jointly around the point input and across the whole image. We note that these features in combination make PointQA a challenging task for existing VQA systems; we have already argued for its importance to human-machine communication.

\smallsec{Need for Visual Grounding} Despite the presence of the point to guide attention, the model must still ground the text in the image (a key challenge in VQA). To verify this, we removed the subject from all questions across the three datasets (e.g. ``What color is \emph{this} shirt?" becomes ``What color is \emph{this}?"), to test if the model indeed benefits from grounding the object word in the image, when given the point input. For PointQA-General, the accuracy (w/ Pythia) drops most significantly from 83.1\% to 75.1\%; the accuracy for PointQA-LookTwice drops from 62.8\% to 56.5\% (Table \ref{table:globalqa-accuracy}), and even for the simpler PointQA-Local setting accuracy drops from 75.0\% to 73.5\%. Moreover, questions in PointQA-General often refer to objects beyond the one directly pointed to, necessitating grounding of those objects in the image (e.g. for the question ``Is this person the farthest from the \emph{yellow box}?", the `yellow box' must be grounded in the image). In a small sample of 100 questions from PointQA-General more than half (55) have this characteristic. Thus visual grounding is a necessary feature of the PointQA task.

\smallsec{Reasoning Beyond the Point Location} For an approximate comparison of accuracy on the three settings we limit each setting to the two most common answers (`white and 'black' for PointQA-Local, '1' and '2' for PointQA-LookTwice) and standardize the model (Pythia three-stream, Sec. \ref{sec:looktwice-models}). PointQA-Local expectedly has the highest accuracy at 90.7\%, while PointQA-LookTwice has a lower accuracy of 68.1\%. This gap partially suggests a key challenge of point-input questions is reasoning about the direct point location \emph{jointly with} the surrounding image context (assessed in the counting questions in PointQA-LookTwice). This is further evidenced by examining those questions in PointQA-General which similarly require contextual reasoning. Accuracy across the comparison words we consider in Sec. \ref{sec:generalqa-results} is 82.2\%, lower than the overall accuracy of 83.1\%. Thus reasoning beyond the immediate point location using the image context is another key challenge of the PointQA task, unique from standard VQA.

\section{Conclusion} 

%\smallsec{Conclusion} 
In summary, we introduced three novel types of visual questions \emph{requiring} a spatial point input for disambiguation. For each question type we created a benchmark dataset and model design and performed extensive analysis. We hope our work inspires further research in this space of questions.

\section{Acknowledgements}

This work is partially supported by
Samsung and by the Princeton SEAS Project-X Funding
Award. Thank you to Karthik Narasimhan, Zeyu Wang,
Felix Yu, Zhiwei Deng, and Deniz Oktay for helpful discussions and feedback on this work. We would also like to thank Sunnie Kim, Zeyu Wang, Sharon Zhang, Angelina Wang, Nicole Meister, Dora Zhao, Ozge Yalcinkaya, Vikram Ramaswamy, and Anat Kleiman for participating in the human evaluations.

% \olga{try to leave space -- if so, let's add a discussion of how one key failure of this analysis is that it doesn't include human points. we've thought about that. those are very challenging to collect, so what we have so far doesn't have language variations and reduces to a segmentation problem. but we still have this dataset, and it's cool, and this is the future direction of this space. }

%\smallsec{Acknowledgments} This work is partially supported by Samsung and by the Princeton SEAS Project-X Funding Award. Thank you to Karthik Narasimhan, Zeyu Wang, Felix Yu, Zhiwei Deng, and Deniz Oktay for helpful discussions and feedback on this work.

%\section{Acknowledgements}
% Zeyu Wang. Internal reviewers. ...

{\small
\bibliographystyle{ieee_fullname}
\bibliography{egbib}
}

\appendix

\section{Human Evaluations}

We conduct human evaluations of the \localqan, \looktwiceqan, and \generalqa datasets to check quality and ensure the reliability of our data collection pipeline. Our methodology is as follows. We select 100 questions at random from each of the test subsets of the three datasets. For each dataset, we ask three volunteers to answer the 100 questions sampled from that dataset; thus our study consists of a total of 9 participants distributed evenly across the three datasets. We send each participant a Jupyter notebook %(interface shown in Fig. \ref{fig:interface}) 
displaying the set of 100 questions and provide the following instructions for each question:

\begin{displayquote}
Please view the image and the small red point in the image, and answer the question to the best of your ability. The red point indicates what the question is asking about. Type in the number corresponding to the answer. Click [Enter] to move on to the next question.
\end{displayquote}

We received 300 responses per dataset (three annotators each for the 100 questions). We record the overall human accuracy (with the denominator being the 300 responses) and for each incorrect response, the potential reason to understand common failure modes in our datasets. 

%\begin{figure}[t]
%\centering
%\includegraphics[width=0.9\linewidth]{latex/figs/Interface.png}
%\caption{Interface shown to human annotators to collect answers for randomly selected point-input questions across the three datasets.}
%\label{fig:interface}
%\end{figure}

%\begin{figure}[t]
%\centering
%\includegraphics[width=0.9\linewidth]{latex/GeneralPie.png}
%\caption{Interface shown to human annotators to collect answers for randomly selected point-input questions across the three datasets [\textcolor{blue}{Nobline: please add this figure}].}
%\label{fig:interface}
%\end{figure}

% \begin{itemize}
%     \item \localqa: 1: red; 2: purple; 3: white; 4: green; 5: yellow; 6: gray; 7: black; 8: blue; 9: orange; 10: pink; 11: brown. The only possible choices are shown for each question. Please select among them.

%     \item \looktwiceqa: 1 through 7 are the only possible choices.
    
%     \item \generalqa: If you think the answer is no, then type in 0 as above. If you think the answer is yes, then type in 1. 0 and 1 are the only possible choices.
    
% \end{itemize}

\subsection{PointQA-General}
\label{sec:generalhuman}

We begin with the main benchmark we introduce in this paper, the \generalqa Dataset of 319,300 questions across 25,420 images. Human accuracy across the 100-question subset (300 annotator answers) is \textbf{90.7\%}. This high accuracy indicates that our question construction and point generation process yields sensible point-input questions that can be reasonably answered by humans. The accuracy of the three-stream model w/ Pythia (Sec. \ref{sec:looktwice-models}) on this 100-question subset is 82.0\%; the substantial human-machine performance gap indicates that the \generalqa dataset is a challenging benchmark for point-input VQA and there is still scope for further model development in this question space.

For the questions in this subset where human annotators disagreed with our dataset answers, we analyze and categorize the reasons for disagreement. The most common are:
\begin{enumerate}
    \item ``Other reasonable answer'' (4.0\%). Another answer in the dataset - in this case \emph{the} other answer since yes/no questions - was a reasonable choice for the question. These correspond to questions in Visual7W that are challenging for human annotators.
    \item ``Point not on object'' (2.3\%). The location of the point is incorrect and the point does not refer to the object being asked about. The fact that this accounts for a very small fraction of errors confirms that drawing the point at the center of the bounding box is a reasonable approach overall.
    \item ``Wrong attention'' (1.3\%). The human annotator pays attention to the wrong object when answering the question. This sometimes occurs when a bounding box answer that is incorrect in Visual7w refers to another object than is asked about in the question, so the annotator pays attention to the correct bounding box. However again this accounts for a small proportion of errors in the dataset.
\end{enumerate}
%Qualitative examples of the most common errors are shown in Fig. \ref{fig:generalex}.
The remaining disagreements are ``question doesn't make sense'' (1.0\%), no obvious reason (0.3\%) and obstruction (0.3\%).

%\begin{figure}[t]
%\centering
%\includegraphics[width=\linewidth]{latex/figs/GeneralExamples.png}
%\caption{Examples of the three most common failure modes in the \generalqa Dataset.}
%\label{fig:generalex}
%\end{figure}

%Qualitative examples of the top three error types are shown in Figure \ref{fig:generalex}.

%\begin{figure}[t]
%\centering
%\includegraphics[width=0.9\linewidth]{latex/GeneralPie.png}
%\caption{Distribution of failure modes in the \generalqa Dataset. 90.7\% of the time human annotators agree with the answer in the dataset. %Possible failure modes in order of prevalence include (1) choosing another reasonable answer (4.0\%), (2) point creating ambiguity (2.3\%), (3) annotator focusing on wrong object (1.3\%), (4) question is not coherent (1.0\%), (5) point obstructs object (0.3\%), (5) the dataset answer is incorrect (1.0\%), and (6) no obvious reason.
%}
%\label{fig:GeneralPie}

%\vspace{5mm}

\subsection{PointQA-Local}
\label{sec:localhuman}

Human accuracy on the 100 question subset of the \localqa Dataset is 75.7\%; this indicates that our questions can be reasonably answered by human annotators. By comparison the accuracy of the \localqa model (Section \ref{sec:localqa-model}) on this question subset is actually 82.0\%; this can be reasoned by the fact that annotators are presented with a larger set of answers (20) in \localqa and when annotating quickly can often choose answers that are reasonable but `less correct' (as discussed below). By contrast in the more robust \generalqa setting the annotators have fewer answer choices and the human-machine gap is a result of the complexity of the question language and higher-level reasoning involved.

\begin{figure}[t]
\centering
\includegraphics[width=\linewidth]{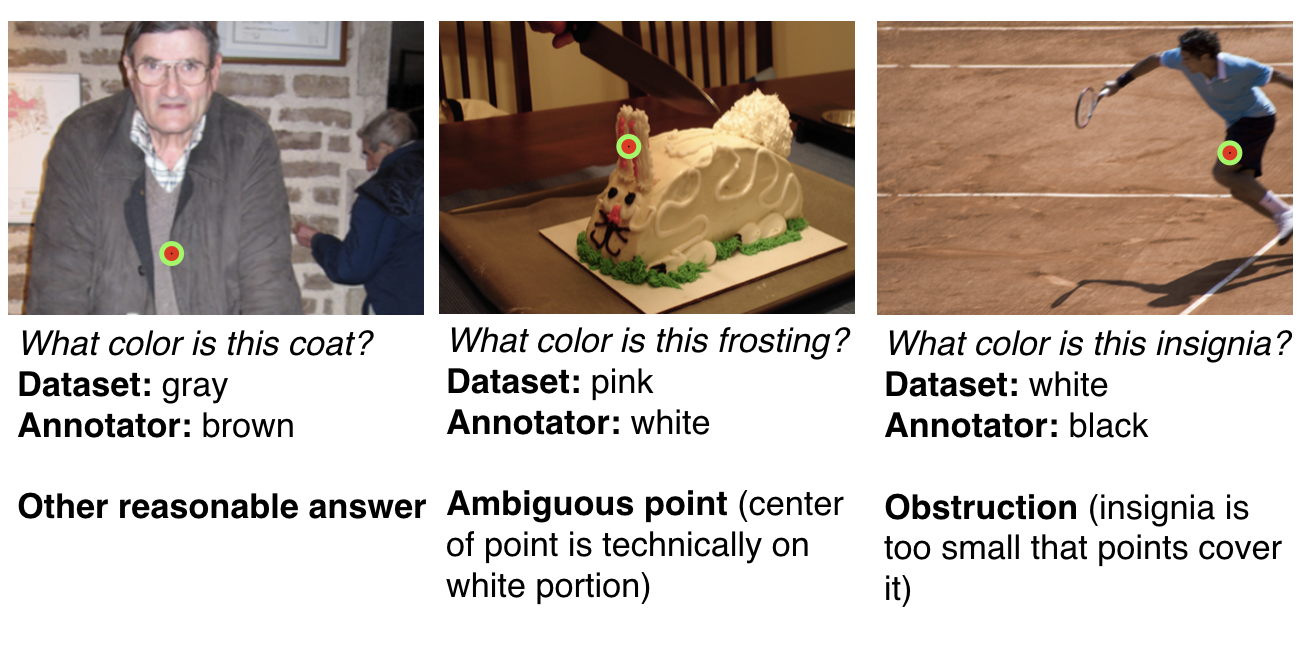}
\caption{Examples of common failure modes in the \localqa Dataset.} %On the left another reasonable answer was chosen by the human annotator; in the middle the annotator focused on the wrong object, and on the right the point created ambiguity.}
\label{fig:localex}
\end{figure}

We further analyze the most common form of errors on the \localqa Dataset. Similar to the \generalqa Dataset, the human errors come from ``other reasonable answer'' (12.3\%), ``ambiguous point'' (4.3\%), ``obstruction'' (3.3\%), ``wrong attention'' (2.3\%), ``answer misannotation (1.0\%) and ``no obvious answer'' (1.0\%). ``Ambiguous point'' refers to cases where it is ambiguous which object is being referred to by the point. ``Obstruction'' refers to cases where the object in question is obstructed by another object in the image or the point. Qualitative examples are shown in Fig. \ref{fig:localex}. The prevalence of the ``other reasonable answer'' category indicates that Visual Genome annotations from which we derived our questions can be accurate but incomplete (for example in Fig. \ref{fig:localex} on the  left, the coat is most accurately labeled gray and not brown, although both labels could be reasonably included in the annotations).

\subsection{PointQA-LookTwice}
\label{sec:lthuman}

\begin{figure}[t!]
\centering
\includegraphics[width=\linewidth]{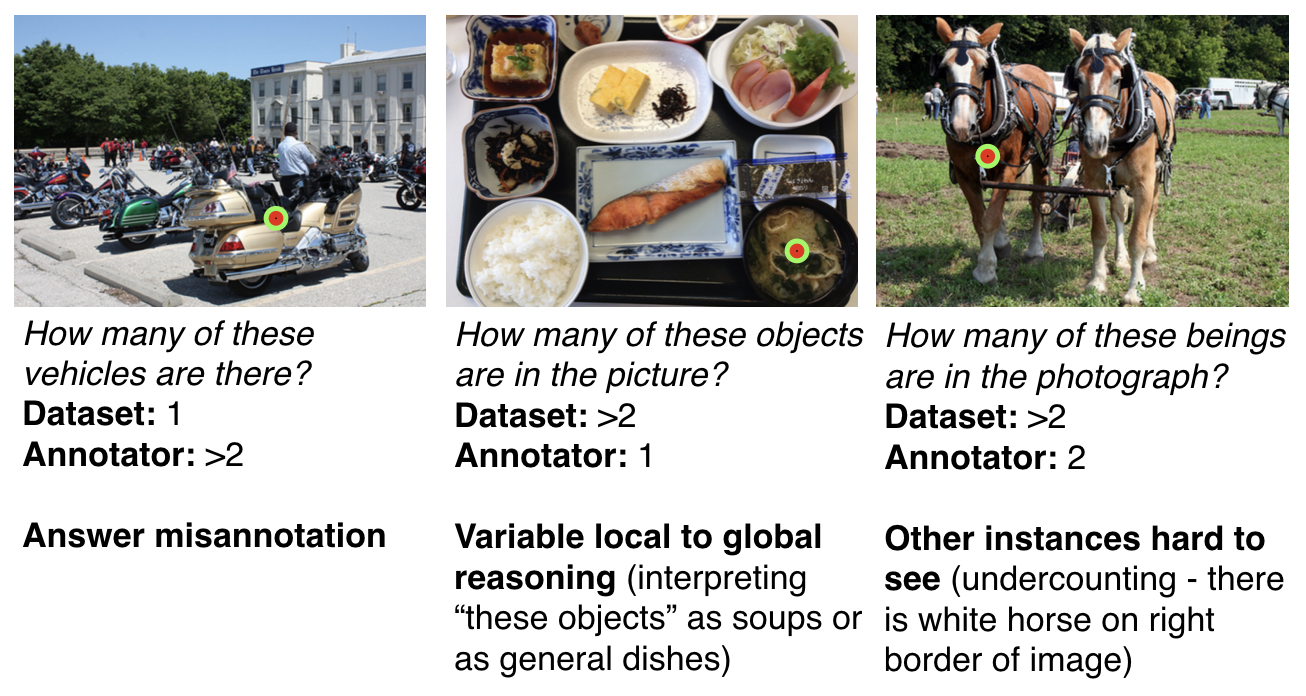}
\caption{Examples of common failure modes in the \looktwiceqa Dataset.} % On the top left the answer is misannotated in the dataset; on the top right other instances of the object in the image are hard to see (white horse in background), and in the middle ``these objects" could be interpreted as referring to the type of food rather than the ``bowl" as intended.}
\label{fig:lookex}
\end{figure}

Human accuracy on the 100-question subset of the \looktwiceqa Dataset is 79.3\%; by comparison the global model of Sec. \ref{sec:looktwice-models} achieves 77.0\%. The most common failure modes are ``answer misannotation'' (7.7\%; i.e., the answer in the dataset is incorrect),  ``variable local-to-global reasoning'' (6.7\%; i.e., the human annotator considered the object pointed to as more generic or specific than intended in the question, and thus over/under-counted),  ``other instances hard to see'' (3.3\%), ``wrong attention'' (1.3\%), ``ambiguous point'' (1.0\%), and ``no obvious reason'' (0.7\%). ``Variable local-to-global reasoning'' is an important failure mode since replacing the object name with a supercategory (e.g. ``these objects") might cause the annotator to count incorrectly; however, the high overall accuracy indicates this is a not a widespread issue.  Qualitative examples are shown in Fig.~\ref{fig:lookex}. In the study, we provided the annotators with possible answer choices of 1 through 7; when evaluating the responses, we narrowed the choices down to 1, 2, $>$2.

Finally, we note that questions were unintentionally drawn from the training set of the \looktwiceqa Dataset. This does not fundamentally affect the quality of our human studies. The model accuracy number of 77.0\% is when excluding these questions from the training set, but could still be inflated. One thing to note is that the source of human disagreement resulting from misannotation in the dataset would be substantially lower without questions from the training set (since the test set consists exclusively of human-written questions).

% \section{Intent-QA Statistics}

% \begin{figure}[!htb]
% \begin{center}
% \includegraphics[width=\linewidth]{latex/will_figures/classwise_distribution.png}
% \end{center}
%   \caption{Response distribution of the Intent-QA Dataset. Responses are of types part (Part), object (Obj), impossible to tell (ITT) and neither object nor part (NA).}
% \label{fig:objpart_class_dist}
% \end{figure}

% \section{Object-Part Example Predictions}

% \begin{figure}[!htb]
% \begin{center}
% \includegraphics[width=\linewidth]{latex/will_figures/binary_object_part_exampls.png}
% \end{center}
%   \caption{Example predictions for object(white) and part (blue). While centroid distance is an important feature, it does not fully capture the task, as shown  in image (a).}
% \label{fig:objpart_class_dist}
% \end{figure}

\end{document}